\documentclass[letterpaper, 10 pt, journal, twoside]{IEEEtran}

\IEEEoverridecommandlockouts                              
\usepackage[T1]{fontenc}
\usepackage{graphicx}
\usepackage{wrapfig}
\usepackage[english]{babel}
\usepackage{amsthm}
\usepackage{amsfonts} 
\usepackage{amsmath}
\usepackage{float}
\usepackage{wrapfig}
\usepackage[table]{xcolor}
\usepackage{tabularx}
\usepackage{siunitx}
\usepackage{subcaption}
\usepackage{enumitem} 
\usepackage{amsmath}
\usepackage{amssymb}
\usepackage{sidecap}
\usepackage{mathtools}
\usepackage{algorithm}
\usepackage{subcaption}
\usepackage{algpseudocode}
\usepackage{booktabs}
\usepackage{cite}
\usepackage{wrapfig}
\usepackage{amsmath}
\usepackage{mwe}
\usepackage{hyperref}
\usepackage{flushend}
\usepackage{arydshln}
\usepackage{multirow}

\usepackage{pifont}

\title{
Learning Causal Structure Distributions\\for Robust Planning
}
\author{Alejandro Murillo-González$^{1}$ and Junhong Xu$^{2}$ and Lantao Liu$^{1}$

\thanks{Manuscript received: February 14th 2025; Revised April 8th 2025; Accepted August 7th 2025. This paper was recommended for publication by Editor Olivier Stasse upon evaluation of the Associate Editor and Reviewers' comments. This work was supported by NSF Award 2047169. } 

\thanks{$^{1}$Alejandro Murillo-González and Lantao Liu are with the School of Informatics, Computing and Engineering, Indiana University, Bloomington, USA {\tt\footnotesize \{almuri, lantao\}@iu.edu}}
\thanks{$^{2}$Junhong Xu is with Nuro, Inc. The work was done while at Indiana University. {\tt\footnotesize juxu@nuro.ai}}
\thanks{Digital Object Identifier (DOI): see top of this page.}
}

\begin{document}

\markboth{IEEE Robotics and Automation Letters. Preprint Version. Accepted August 2025}
{Murillo-González \MakeLowercase{\textit{et al.}}: Learning Causal Structure Distributions for Robust Planning} 

\maketitle

\vspace{-10pt}
\begin{abstract}
Structural causal models describe how the components of a robotic system interact. They provide both structural and functional information about the relationships that are present in the system. The structural information outlines the variables among which there is interaction. The functional information describes how such interactions work, via equations or learned models. In this paper we find that learning the functional relationships while accounting for the uncertainty about the structural information leads to more robust dynamics 
models which improves downstream planning, while using significantly lower computational resources. This in contrast with common model-learning methods that ignore the causal structure and fail to leverage the sparsity of interactions in robotic systems. We achieve this by estimating a causal structure distribution that is used to sample causal graphs that inform the latent-space representations in an encoder-multidecoder probabilistic model. We show that our model can be used to learn the dynamics of a robot, which together with a sampling-based planner can be used to perform new tasks in novel environments, provided an objective function for the new requirement is available. We validate our method using manipulators and mobile robots in both simulation and the real-world. Additionally, we validate the learned dynamics' adaptability and increased robustness to corrupted inputs and changes in the environment, which is highly desirable in challenging real-world robotics scenarios. Video: \url{https://youtu.be/X6k5t7OOnNc}.
\end{abstract}

\begin{IEEEkeywords}
Model Learning for Control; Integrated Planning and Control; Reinforcement Learning; Planning under Uncertainty; Unstructured Environment.
\end{IEEEkeywords}

\vspace{-5pt}
\section{Introduction}

\IEEEPARstart{S}{tructural} Causal Models (SCM) can explain how the components of a robotic system interact \cite{pearl2009causality}.
A significant advantage of explicitly modeling structured causal relationships compared to non-structured approaches is enhanced generalization, resulting in greater robustness to varying test conditions \cite{scholkopf2021toward}.
This improved generalization stems from the inherently sparse interactions among components in most robotic systems.
For instance, an actuator typically only influences the joints directly connected to it.
Consequently, even if one component's input is corrupted a SCM can still provide accurate predictions for other components.

Combined with model-based planners~\cite{botev2013cross, gandhi2021robust}, this robustness property can lead to reliable planning performance.
However, in most cases, the true SCM is unknown, requiring it to be estimated from data or hand-designed by an expert~\cite{scholkopf2021toward}.
This estimation presents two significant challenges.
First, it is computationally expensive due to the combinatorial growth in possible causal structures as the number of states and control variables increases \cite{robinson1977counting}.
Second, even with extensive data, the inherent observational equivalence of multiple causal structures makes it virtually impossible to identify a single, uniquely correct structure~\cite{pearl2009causality}.

Given the potential of SCMs in robotics (e.g., \cite{cannizzaro2023car, ding2023causalaf, wang2022causal}) but recognizing the inherent difficulties in their estimation, this paper proposes a novel approach to address these challenges.
Our key finding is that \textit{learning a distribution over the causal structures for a robot's dynamics, instead of a single causal graph, gives downstream planners improved robustness}
to common challenges faced by robotic systems.
For this, we introduce two core ideas. First, we can constrain the SCM search space by recognizing that the dynamics of the robot can only flow forward in time from the current state and action to the next state.  
This allows us to focus on searching bipartite causal graphs with directed nodes from state and action variables, conforming to the Markovian property.  
Second, we find that we can use the Integrated Gradients \cite{integrated_gradients} feature attribution method to efficiently estimate the parameters of the dynamics' causal structure distribution.

\begin{figure}
\begin{center}
\includegraphics[width=0.65\linewidth]{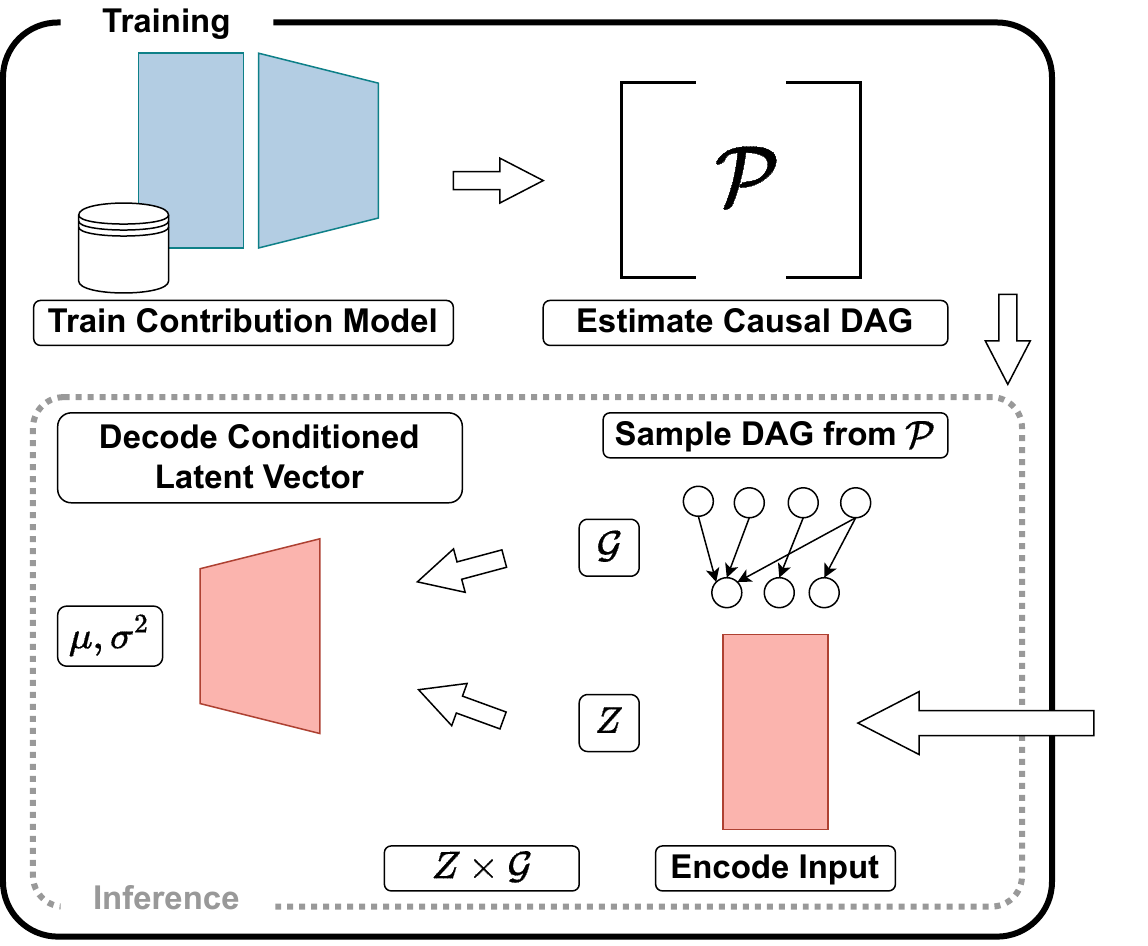}
\centering
\caption{
\small{
Proposed CAusally-informed DYnamics (CADY) model. We train a contribution model $f^C$ using the robot's experience to estimate a distribution over causal structures $\mathcal{P}(\mathbf{p})$. A probabilistic encoder-multidecoder $f^D$ (inside the dotted box) masks its latent space using sampled causal structures from $\mathcal{P}(\mathbf{p})$, leading to a dynamics model that improves downstream planning robustness. 
}\vspace{-20pt}
}
\label{fig:overview}
\end{center}
\end{figure}
In addition to improving the dynamics estimation process, we show that learning the functional relationships while accounting for the uncertainty about the structural information, via the estimated causal structure distributions,
results in more robust models.  
Specifically, these models are better equipped to handle unmodeled interventions, changes in the system, and challenges such as missing or noisy data.
To demonstrate this, we introduce a {\em CAusally-informed DYnamics model (CADY)}, based on a probabilistic encoder-multidecoder architecture that uses causal structures sampled from the estimated distribution to mask its latent representations. 
By causally-informed/inspired, we mean that our approach draws ideas from causal theory without strictly adhering to it. In our case, we do it by using estimated causal relationships to improve predictive accuracy and robustness. \cite{pearl1995causal, huang2022adarl, feng2023learning}.
Fig. \ref{fig:overview} provides an overview.

Overall, the contributions of this paper include:
\begin{itemize}[leftmargin=0.4cm]
\item
We introduce a new framework based on feature attribution to efficiently estimate the \textit{distribution of causal structures of a robot's dynamics}, avoiding expensive tests for causal discovery while yielding a distribution of structures rather than a single causal graph.

\item
We present CADY, \textit{a causally-informed model with greater robustness to real-world challenges} such as inaccurate state estimates, missing sensor data, and dynamics changes, thereby enabling more reliable robot planning. 

\item
Via extensive experiments, we show that  
CADY outperforms existing approaches while also using fewer computational resources, 
as we demonstrate in manipulation and navigation tasks --in simulation and real-world-- using the learned causally-informed dynamics. 
\end{itemize}

\vspace{-4mm}
\section{Related Work}

\vspace{-1mm}
We review methods for causal structure learning, noting the literature's focus on causal graphs over distributions, and then discuss robot dynamics learning.

\textbf{Causal Structure Learning.} In \cite{saeed2020causal} distributions of causal structures are learned by assembling a set of $K$ causal graphs on the same set of nodes. 
Unlike our approach, their method assumes that the $K$ graphs are already provided, and therefore does not address the problem of identifying these causal relationships.
Other causal discovery methods include \cite{spirtes1991algorithm, lorch2022amortized, bello2022dagma, wu2024nonlinear},
which rely on multiple strategies to determine the Bayesian network describing the causal relationships. For example, statistical independence tests, optimization constraints or via a scoring function. However, in general, these are only interested in obtaining a single causal structure and do not characterize the uncertainty over the causal relationships as we do. Applications include: out-of-distribution generalization, data augmentation, robust MPD solution \cite{liu2021learning, lu2021invariant, he2022causpref, pitis2020counterfactual, ding2023seeing, pitis2022mocoda}. 

\textbf{Dynamics Learning.} To date, robot dynamics learning has been seen as a task- and/or embodiment-specific problem 
\cite{jiahao2023online,wang2017dynamic,wang2021dynamics,wang2022causal,hartmann2013real,rezaei2019cascaded} which however limits its generalization to a broader scope of applications.
Another option is dynamic systems theory, which manually models and analyzes how variables interact and evolve over time \cite{luenberger1979introduction,khosla1985parameter}. These methods focus on the relationships between variables, their rates of change, and the feedback loops that drive the system's behavior. An example is a set of differential equations describing the evolution of a system, obtained by an expert in a difficult, time consuming process,  which are not straightforward to adapt to changes in the set of assumptions used to derive them \cite{mh1974differential}. On the other hand, causal structure learning aims to infer causal relationships by leveraging various assumptions and statistical techniques. They often rely on interventions or experiments to study how changes in one variable affect others \cite{deleu2022bayesian,lorch2022amortized,nishikawa2022bayesian}. Specific to robotics, some causal-based methods have been proposed for problems such as failure explanation \cite{diehl2022did}, driving scenario generation \cite{ding2023causalaf} and manipulation \cite{wang2022causal}.


\vspace{-2mm}
\section{Preliminaries}
\vspace{-1mm}
\subsection{Causal Structure Learning} \label{subsec:prelim-causal}
\vspace{-1mm}

\textbf{Structural Causal Models (SCM)} describe the variables used to represent a system and the causal-effect relationships among them~\cite{pearl2009causality, scholkopf2021toward}. The variables are divided in two non-empty finite sets, $A$ and $B$. The set of \textit{exogenous variables}, $A$, are those whose values are provided to the model without 
requiring an explanation. 
The set of \textit{endogenous variables}, $B$, consists of variables that we want to understand how they depend and change according to a subset of the remaining variables in the system (this includes both exogenous and endogenous variables). For each endogenous variable $b \in B$ there is a function $f_b : W \rightarrow dom(b)$, where $W \subseteq (A \cup B) \setminus \{b\}$ and $dom(b)$ is $b$'s {\em domain} represented by a non-empty set of values that $b$ can take. 
The function $f_b$ describes how the value of $b$ depends on each variable $w \in W$. Each variable in $W$ is called a \textit{direct cause} of $b$, and a variable is a \textit{cause} of $b$ if it is a direct cause of $b$ or an indirect cause of any variable in $W$. At least one of the causes of $b$ must be an exogenous variable. 
All  SCMs are {graphically represented} via a \textit{directed acyclic graph} (DAG) $\mathcal{G} = (\mathcal{N}, \mathcal{E})$ where $\mathcal{N} = A \cup B$ is the set of nodes and $\mathcal{E}$ the set of directed edges. 
If node $n$ is a direct cause of node $n^{\prime}$ ($n \neq n^{\prime}$), then there is a directed edge going from $n$ into $n^{\prime}$. 

\textbf{Causal Structure Learning} aims to discover --using data-- the SCM that describes how a system evolves over time. This is similar to how differential equations model the evolution of a system, but {\em without requiring the expert knowledge to formulate them}, which is a difficult and time-consuming process prone to failure when the differential equations fail to model challenging physical phenomena. 
On the other hand, methods driven purely by correlation-based learning usually assume the availability of large amounts of \textit{i.i.d.} data, 
but they fail when faced with distribution shifts in the input data \cite{scholkopf2021toward}. 

In contrast, we strengthen the robustness of the dynamics model by estimating and integrating the system's causal relationships and their associated uncertainties. 

\vspace{-3mm}
\subsection{Integrated Gradients for Feature Attribution}
\vspace{-1mm}
Feature Attribution (FA) involves determining how each element of the input of a deep neural network contributed to its final prediction \cite{zhou2022feature, nguyen2021effectiveness}. In this work, we employ FA to estimate the parameters of the probability distribution used to model the causal structures, which we present in Section \ref{sec:scm-dist}. In particular, we use {\em Integrated Gradients (IG)} which was introduced in \cite{integrated_gradients} as a FA method.

In essence, 
 IG takes two inputs. The first one, $\mathbf{x}$, is the value for which we want to carry out FA. The second one, $\mathbf{x}^\prime$, which we called the \textit{null input}, is used as a reference input with respect to which FA takes place. 
Specifically, the \textit{null input} $\mathbf{x}^\prime$ is chosen so that whenever it is fed to a model $\mathcal{M}$ the output from the model is approximately zero. Formally, \vspace{-2mm}
\begin{align*}
&\text{Let}~\mathbf{x}^\prime \in \mathbb{R}^d~\text{be an input to the neural network}~\mathcal{M}.\\
&\text{We call}~\mathbf{x}^\prime~\text{a \textit{null input} of}~\mathcal{M}~\text{if}~\mathcal{M}(\mathbf{x}^\prime) \approx \boldsymbol{0}.
\end{align*} 
Using the \textit{null input}, IG computes the ``path integral of the gradients along the straight-line path from the baseline [\textit{null input}] to the \textit{input}'' \cite{integrated_gradients}, where \textit{input} refers to the vector $\mathbf{x} \in \mathbb{R}^d$ whose FA we aim to analyze. This process allows IG to assess the contribution of each input variable to the model’s output. 
Specifically, IG produces an attribution vector $\mathbf{a} \in \mathbb{R}^d$, defined as $\text{IG}(\mathbf{x}, \mathbf{x}^\prime; \mathcal{M}) = \mathbf{a}$
where each element $a_i$ of $\mathbf{a}$ quantifies the contribution of the $i$-th feature in the input $\mathbf{x}$ to the prediction made by the model $\mathcal{M}$. A larger magnitude of $a_i$ indicates a stronger influence of the $i$-th input feature on $\mathcal{M}$'s output.


\vspace{-2mm}
\section{Method} \label{sec:Method}
\vspace{-1mm}

We propose to learn dynamics models that are aware of the uncertainty over its inputs. We show that we can achieve this by modelling the distribution $\mathcal{P}$ over the causal structures that best explains the relationships between the observed data. We then use $\mathcal{P}$ to generate feasible causal structures used to mask the latent space of an encoder-multidecoder probabilistic dynamics model. 
\vspace{-4mm}
\subsection{Learning the SCM Distribution} \label{sec:scm-dist}
\vspace{-1mm}
Due to the complexity of the real world, the collected data may exhibit bias originated from observational data~\cite{xu2022causal}, be contaminated by extraneous noises~\cite{glymour2019review}, or exclude necessary causal variables~\cite{zhang2020causal}. 
As a result, multiple causal graphs could potentially explain the same dataset \cite[Chapter~2.3]{pearl2009causality}, 
each exhibiting similar likelihoods of representing the data.
Thus, \textit{relying solely on a single plausible causal structure risks overlooking the potentially correct structure}, which may lead to inaccurate predictions of interventions (control actions) and, subsequently, unsuccessful planning and decision-making~\cite[Chapter 4]{pearl2009causality}.
To address this, we propose to learn a {\em distribution} over possible causal graphs, allowing downstream planners to consider multiple plausible causal structures when predicting an action's effects, thus enhancing the overall robustness of planning outcomes.

\textbf{Bounding the SCM Search Space.} \label{sec:bounded_search_space}
In \cite{robinson1977counting} it was shown that the number of DAGs that can be constructed using $n$ nodes is given by: $a_n = \sum_{\substack{k=1}}^{\substack{n}} (-1)^{k-1} \binom{n}{k} 2^{k (n-k)}a_{n-k}$.
Therefore, as the number of state variables or control inputs in a system increase, the potential causal relationships and graph structures expand combinatorially. This leads to an enormous search space that quickly becomes computationally infeasible to explore exhaustively. 

\begin{wrapfigure}{r}{0.6\linewidth}
    \centering
    \includegraphics[width=\linewidth]{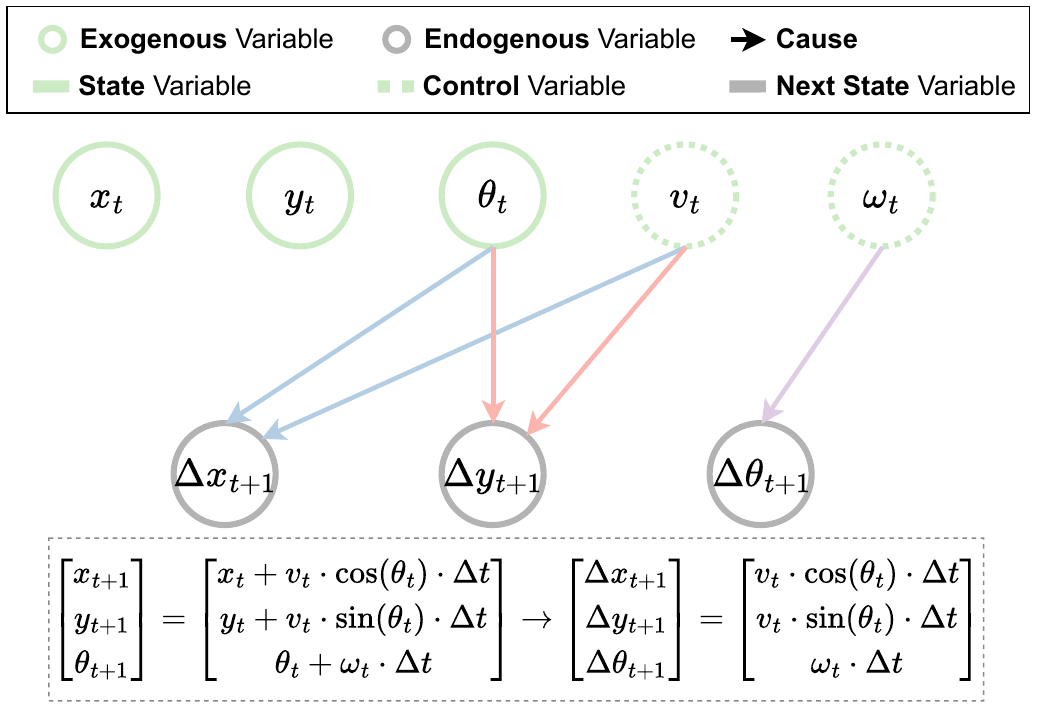}
    \caption{\small \textit{Dynamics} causal graph  
    with $n=3$ state ($[x_t, y_t, \theta_t]$) and $p=2$ control ($[v_t, \omega_t]$) variables for a discretized differential drive robot model \cite{dubins-dyns}. For clarity, functional relationships are shown in the dotted box, and the discretization constant $\Delta t$ is omitted. \vspace{-10pt}
    }
    \label{fig:causal-graph-example}
\end{wrapfigure}

However, we identify a pattern in the dynamics modeling task that we can exploit to reduce the SCM search space. Formally, consider the state vector \(\mathbf{s}_t \in \mathbb{R}^n\) and action variables \(\mathbf{a}_t \in \mathbb{R}^p\) at time \(t\). The only edges in our causal graph must be those from \(\mathbf{s}_t\) and \(\mathbf{a}_t\) to \(\mathbf{s}_{t+1}\), since nodes at time $t$ cannot influence each other and nodes at $t+1$ cannot influence something in the past. This forms a {\em bipartite DAG} where the nodes representing \(\mathbf{s}_{t+1}\) have incoming edges \textit{only} from nodes representing \(\mathbf{s}_t\) and \(\mathbf{a}_t\). 
Fig. \ref{fig:causal-graph-example} presents an example. 
\textit{Consequently, we will focus on constructing such bipartite DAGs 
which will substantially reduce the search space of SCMs for the robot's dynamics.}

\textbf{Modelling the SCM Distribution.} \label{sec:scm_dist_model}
We propose to learn the distribution over causal structures of the robot's dynamics by modelling the distribution \(\mathcal{P}\) of the edges in a causal graph \(\mathcal{G}\). This distribution consists of \(n + p\) exogenous and \(n\) endogenous nodes (which can be interpreted as the scale of the causal graph). See the example in Fig.~\ref{fig:causal-graph-example}. 

Let \(E_{ij}\) be a random variable representing the existence of an edge from node \(i \in \{1, \dots, n+p\}\) (representing state or actions at time $t$) to node \(j \in \{1,\dots,n\}\) (representing state at time $t+1$) in the causal graph \(\mathcal{G}\). Each \(E_{ij}\) follows a Bernoulli distribution with parameter \(p_{ij}\) (interpreted as probability of an edge from node $i$ to node $j$). Thus, the probability mass function (PMF) of \(E_{ij}\) is given by:
\begin{equation} \vspace{-1mm}
\Pr(E_{ij} = e_{ij}) = p_{ij}^{e_{ij}} (1 - p_{ij})^{1 - e_{ij}}, \quad \text{for } e_{ij} \in \{0, 1\}. 
\end{equation}
Since the existence of an edge from one node to another is independent of the existence of edges among other non-related variables, we have that the variables \(E_{ij}\) are independent and the joint probability distribution \(\mathcal{P}\) is the product of the individual Bernoulli distributions. The joint PMF of the set \(\{E_{ij}\}\) is:
\begin{equation} \vspace{-3mm}
    \Pr(\{E_{ij} = e_{ij}\}; \mathbf{p}) = \prod_{i,j} p_{ij}^{e_{ij}} (1 - p_{ij})^{1 - e_{ij}}, \label{eq:distribution_def}
\end{equation}
where \(\mathbf{p} = (p_{ij})\) is the matrix of Bernoulli parameters.

\textbf{Estimating the SCM Distribution Parameters $\mathbf{p}$.}
To avoid a computationally expensive search for causal relationships, we employ the IG method to approximate the contribution of each input feature to the system's output. Although IG was not designed as a causality test, we use it to estimate the likelihood of each feature being a cause in the underlying SCM, as comparing against a \textit{null input} $\mathbf{x}^\prime$ (which is the key in the IG method) involves implicitly ``assign(ing) blame to a certain cause'' \cite{integrated_gradients}. That is, causal influence can be measured from the effect of “perturbing” a parent on its child while holding others fixed. IG’s construction—integrating gradients from a null input—provides an efficient approximation of that effect without resorting to combinatorial graph search.

Using IG, we construct the \((n + p) \times n\) matrix \( \mathbf{p} \) from Eq. \eqref{eq:distribution_def}. We estimate each element \( {p}_{ij} \in \mathbf{p} \) as:
\begin{align}
    p_{ij} &\coloneqq \text{IG}_{ij}(\mathbf{x}, \mathbf{x}'; \mathcal{M}) \label{eq:ig2p} \\ 
    &= (x_i - x_i^\prime) \cdot \int_{\alpha = 0}^1 \frac{\partial}{\partial x_i} \mathcal{M}(\mathbf{x}^\prime + \alpha \times (\mathbf{x} - \mathbf{x}^\prime))_j \, d\alpha, \nonumber 
\end{align}
where we set the \textit{null input} $\mathbf{x}' = \mathbf{0}$ and $\mathcal{M}$ is the model used for analyzing the contribution. 

We then interpret each \(p_{ij}\) as a score proportional to the (unnormalized) probability that feature \(x^i_t \vee a^i_t \) is as a cause of output $x_{t+1}^j$.  Concretely, we normalize across all candidate parents to obtain the Bernoulli parameters \(\mathbf{p} = (p_{ij})\) for the distribution \(\mathcal{P}\):
\begin{equation}
    p_{ij} = \max\left(\rho_{min}, \min\left(\frac{\mathtt{s}(|p_{i,j}|)}{\max_{j} \mathtt{s}(|p_{:,j}|)}, 1 - \rho_{min} \right)\right),    \label{eq:p_processing}
\end{equation}
where $\mathtt{s}(\cdot)$ is an optional smoothing function to avoid one contribution from significantly out-weighting the others and $0 < \rho_{min} < 0.5$ clips the minimum and maximum probability, which we empirically found improved performance as it regulates the model's certainty about its estimations. 

\vspace{-2mm}
\subsection{CAusally-informed DYnamics Model (CADY)} \label{sec:scm-dist-plan}
\vspace{-1mm}

\textbf{Model Architecture.}
To model the underlying dynamics, we propose to use a probabilistic neural network with an encoder-\textit{multi}decoder architecture, whose latent space is masked using causal structures sampled from the causal structure distribution $\mathcal{P}$ defined in Eq. \eqref{eq:distribution_def}. Our dynamics model (CADY) is defined as:
\begin{equation}
    f(\mathbf{x}) = f_{dec}(f_{enc}(\mathbf{x}) \odot \mathbf{M}), \quad \mathbf{M} \sim \mathcal{P}(\mathbf{p}),  \label{eq:cady}
\end{equation} which encompasses:

\begin{itemize}[leftmargin=0.4cm]

    \item \textit{Encoder:} Let \(\mathbf{s} \in \mathbb{R}^n\) be the vector of state variables and \(\mathbf{a} \in \mathbb{R}^p\) be the vector of action variables. Define the concatenated input vector:
            $ 
                \mathbf{x} = [\mathbf{s}; \mathbf{a}] \in \mathbb{R}^{n+p}.
            $ 
            The encoder \(f_{\text{enc}}: \mathbb{R}^{n+p} \to \mathbb{R}^{n+p}\) maps this input into a latent space of the same dimension:
            $ 
                \mathbf{z} = f_{\text{enc}}(\mathbf{x}) \in \mathbb{R}^{n+p}.   \label{eq:latent-var}
            $ 

    \item \textit{Latent Vector Masking:}
            Let \(\mathbf{M} \in \{0, 1\}^{(n+p) \times n}\) be the mask matrix sampled from the learned distribution \(\mathcal{P}\). Each element \(M_{ij}\) indicates the presence or absence of an edge from the \(i\)-th latent feature to the \(j\)-th next state variable. Define the masking process as:
            \begin{equation}
                \mathbf{z}_j = \mathbf{z} \odot \mathbf{m}_{:,j} \quad \text{for } j = 1, 2, \ldots, n,
            \end{equation}
            where \(\mathbf{m}_{:,j}\) is the \(j\)-th column of \(\mathbf{M}\) and \(\odot\) denotes element-wise multiplication. By masking we select the latent features to use for predicting each next state variable. Note that $\mathbf{M} \sim \mathcal{P}(\mathbf{p})$ is sampled every time $f(\cdot)$ is called.

    \item \textit{Decoder:} Each masked vector \(\mathbf{z}_j\) is fed to the corresponding decoder \(f_{\text{dec}}^j: \mathbb{R}^{n+p} \to \mathbb{R}^2\), which outputs the parameters of a Gaussian distribution for the next
                state variable: 
            $ 
                (\mu_j, \sigma_j^2) = f_{\text{dec}}^j(\mathbf{z}_j).
            $ 
            The 
            next state variable \(s_j'\) is then:
            $ 
                s_j' \sim \mathcal{N}(\mu_j, \sigma_j^2).
            $ 
            This way we account for the aleatoric uncertainty due to unexpected and un-modeled disturbances into the prediction\cite{pets}.

\end{itemize}

\textbf{Training \& Inference.}
During learning we instantiate two CADY models (Eq.~\ref{eq:cady}): one for causally-informed dynamics prediction, denoted as \( f^D \), and another, the contribution model \( f^C \), which will be used to estimate the distribution \( \mathcal{P}(\mathbf{p}^D) \), from which \( f^D \) samples causal masks. By connecting every input feature \(x_i\) to every hidden neuron (and ultimately every output \(x_{t+1}^j\), \(f^C\) ensures that no potential causal link is precluded by architectural sparsity.  This ``fully wired" design is essential: if we omitted any connection in \(f^C\), we could not attribute an input–output pair, biasing the learned causal‐structure distribution. Specifically, for \( f^C \), we constrain all elements of the parameter vector \( \mathbf{p}^C \) to one, i.e.:  $\quad p^C_{ij} = 1, \quad \forall i, j.$
This configuration ensures that all input variables to \( f^C \) remain active during latent vector masking, thereby simulating a scenario in which all inputs are treated as causes of all output variables.

The training procedure follows a sequential approach. First, we train \( f^C \) using the available dataset $\mathcal{D}$. Once trained, we employ \( f^C \) as the model IG will use to compute an estimate of \( \mathbf{p}^D \) via Eq.~\eqref{eq:ig2p} and subsequently transforming each element \( p_{ij} \in \mathbf{p}^D \) through Eq.~\eqref{eq:p_processing}. The resulting parameter vector \( \mathbf{p}^D \) then defines the distribution from which the causal masks \( \mathbf{M} \) are sampled during the training of \( f^D \) with the data in $\mathcal{D}$.

At inference time, \( f^C \) is discarded, and the final learned value of \( \mathbf{p}^D \) is employed to parameterize the causal mask distribution \( \mathcal{P}(\mathbf{p}^D) \), ensuring that the causal relationships inferred during training are preserved in deployment. Fig. \ref{fig:overview} shows the training and inference processes.
 
\vspace{-7pt}
\subsection{Planning with CADY} 
\vspace{-1mm}

The action selection process follows a Model Predictive Control (MPC) strategy, where \( f^D \) is integrated with a sampling-based planner to iteratively optimize a given objective by selecting the most advantageous actions. Planning over the distribution of causal structures is achieved by masking the latent variable \( \mathbf{z} \) at each timestep, as \( f^D \) utilizes randomly sampled causal structures \( \mathbf{M} \) drawn from \( \mathcal{P}(\mathbf{p}^D) \). This approach allows the planner to leverage learned causal dependencies while accounting for epistemic uncertainty arising from structural ambiguity. Furthermore, aleatoric uncertainty is naturally incorporated through the probabilistic nature of \( f^D \), which parameterizes a Gaussian distribution over predicted outcomes. This dual treatment of uncertainty enhances the robustness of the planning process, leading to more reliable action selection \cite{hofer2002approximate, pets}.

\vspace{-2mm}
\section{Results}
\vspace{-1mm}

Our experiments are designed to assess three critical aspects:  
\textbf{(Q1)} Can CADY reduce computational demands, enabling efficient onboard inference without sacrificing accuracy?  
\textbf{(Q2)} Does CADY exhibit greater resilience to noisy and missing state estimates?  
\textbf{(Q3)} How well does CADY adapt to changes in the dynamics regime?  

To address \textbf{(Q1)}, we evaluate learning efficiency and computational requirements in Sections \ref{subsec:results.learning} and \ref{subsec:results.comp-req}, respectively. For \textbf{(Q2)}, we assess robustness by introducing varying levels of noise and partial state observability (missing values) in Section \ref{subsec:results.robustness}, demonstrating CADY’s superior reliability in real-world scenarios, thus enabling more robust planning. Section \ref{subsec:results.adaptation} examines the model's ability to adapt to unexpected shifts in robot dynamics, addressing \textbf{(Q3)}. Finally, Section \ref{subsec:results.field-exp} presents results from our field experiments, providing real-world validation of CADY’s capabilities.  

\vspace{-2mm}
\subsection{Evaluation Setup} \label{subsec:eval-setup}
\vspace{-1mm}

\textbf{Baselines.} We consider \textit{six} competitive methods spanning \textit{three complementary categories}: (i) \textit{causally-inspired dynamics models} — CMI \cite{wang2022causal}, which infers a causal graph via conditional mutual information but requires manual threshold selection to distinguish dependencies, and REG \cite{wang2021task}, which induces sparse causal and intervention masks via L1 regularization to abstract state and action inputs for downstream planning; (ii) \textit{model-based learners} — PETS \cite{pets}, a probabilistic ensemble that achieves sample-efficient, state-space dynamics learning for continuous control; and (iii) \textit{model-free RL} — PPO \cite{schulman2017proximal}, SAC \cite{haarnoja2018soft}, and TD3 \cite{fujimoto2018addressing}, chosen as data-driven methods with strong empirical performance. 
Model-free baselines are implemented using Stable-Baselines3 \cite{stable-baselines3}, while others use the respective authors' code. 
We make no assumptions about the robot encountering specific challenges and do not incorporate such information during training. Consequently, we exclude methods that explicitly correct for inaccuracies, such as physics models with Kalman-filtered state estimates. 
However, CADY remains compatible with such enhancements, and their use further mitigates the impact of noise.  

\begin{figure} 
    \centering
    \includegraphics[width=0.28\linewidth]{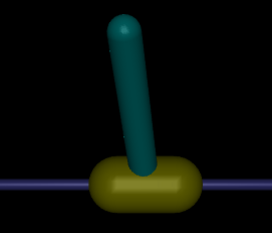}
    \includegraphics[height=0.24\linewidth]{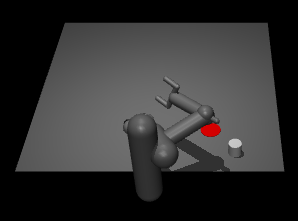}
    \includegraphics[width=0.275\linewidth]{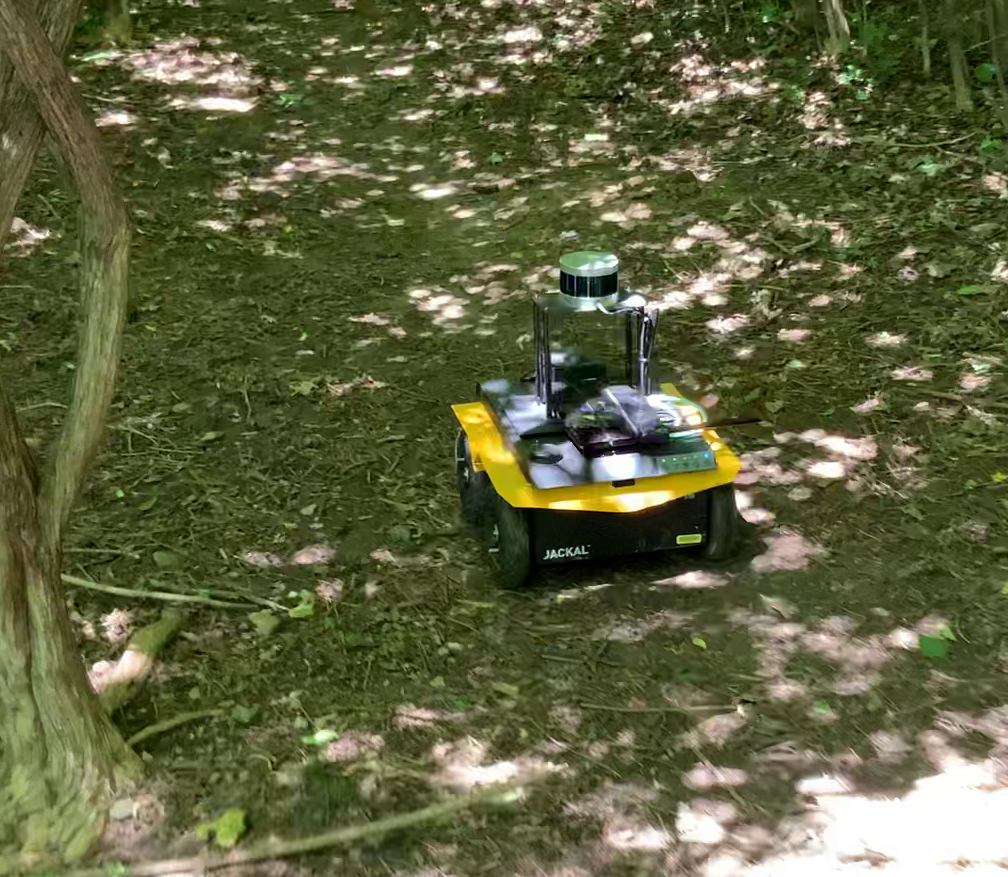}
    \caption{\small{Evaluation Platforms. \textit{(Left)} Cartpole. \textit{(Center)} Manipulator. \textit{(Right)} Ground vehicle tested in simulation and the real-world.}\vspace{-20pt}}
    \label{fig:evaluation-platforms}
\end{figure}

\textbf{Platforms and Environments.}  Our evaluations include manipulators and mobile robots in simulation \cite{mbrl-lib, inspection_world} and Clearpath's Jackal in the real world (see Fig. \ref{fig:evaluation-platforms}):

\begin{itemize}[leftmargin=0.4cm]
    \item \textit{Continuous Cartpole}: Balance a pole on a moving cart with state $[x_t, \theta_t, \dot{x}_t, \dot{\theta}_t]^T \in \mathbb{R}^4$ and action $a \in [-3, 3]$.  
    \item \textit{Pusher}: Robotic arm moving an object to a target. The state $\mathbf{x} \in \mathbb{R}^{20}$ includes joint and end-effector positions, with actions $\mathbf{a} \in [-2, 2]^7$ representing applied torques.  
    \item \textit{Jackal}: Ground vehicle with state $[x_t, y_t, \theta_t]^T \in \mathbb{R}^3$, controlled via linear velocity $v \in [-1, 1]$ and angular velocity $\omega \in [-\pi/3, \pi/3]$. We test on uneven (Inspection World \cite{inspection_world}) and even terrains (Gazebo empty world).  
\end{itemize}

\textbf{Implementation Details.} CADY follows $f^D(\mathbf{x}) = f_{dec}(f_{enc}(\mathbf{x}) \odot \mathbf{M})$, where $f_{enc}$ is a linear layer with $(n + p)$ hidden units, and each output dimension has a 3-layer decoder $f_{dec}$ with hidden sizes 3, 13, and 20 for Cartpole, Pusher, and Jackal, respectively. Outputs include per-dimension mean and log-variance. We normalize inputs/outputs and learn the state delta $\Delta \mathbf{s}_{t}$ ($\mathbf{s}_{t+1} = \mathbf{s}_t + \Delta \mathbf{s}_t$). The networks are randomly initialized, and variance is bounded to prevent collapse \cite[Appendix A.1]{pets}. Parameters of $\mathbf{p}^D$ start at $0.5$, with $\mathtt{s}(\cdot)$ set to the cubic root and bound $\rho_{min} = 0.02$. 
A batch size of 8 enforces frequent causal graph re-sampling. Training runs for up to 64 epochs or until loss change is below $10^{-3}$. Optimization uses Negative Log-Likelihood (NLL) loss: $\mathcal{L}(\theta) = - \log \hat{\sigma}^2 ||\hat{\mu}, y||_2 + \log \hat{\sigma}^2$, with \textit{Adam} (learning rate $3\text{e-}3$). $f^C$ is a copy of $f^D$, but $\mathbf{p}^C$ is fixed as a matrix of ones as described in Section \ref{sec:scm-dist-plan}.  

For planning, we use Cross-Entropy Method (CEM) \cite{botev2013cross} in Cartpole and Pusher, with re-plan frequency 1, $\alpha=0.1$, and elite ratio 0.1. Jackal uses Model Predictive Path Integral (MPPI) \cite{gandhi2021robust} with $\gamma=0.9$, $\sigma=0.01$, and $\beta=0.6$.  

All experiments were performed on an Intel NUC (Core i7-1165G7, NVIDIA RTX 2060 GPU) with Ubuntu 20.04.

\vspace{-5mm}
\subsection{Learned Models} \label{subsec:results.learning}
\vspace{-1mm}

We begin by validating that CADY supports different dynamics learning strategies. In particular, we learn the models for Cartpole and Pusher in a Model-Based Reinforcement Learning (MBRL) fashion. For Jackal, we do system identification with a pre-collected simulation dataset, which we find to be sufficient for real-world deployment. 

\begin{figure}
    \centering
    \includegraphics[width=0.47\linewidth, height=2.2cm]{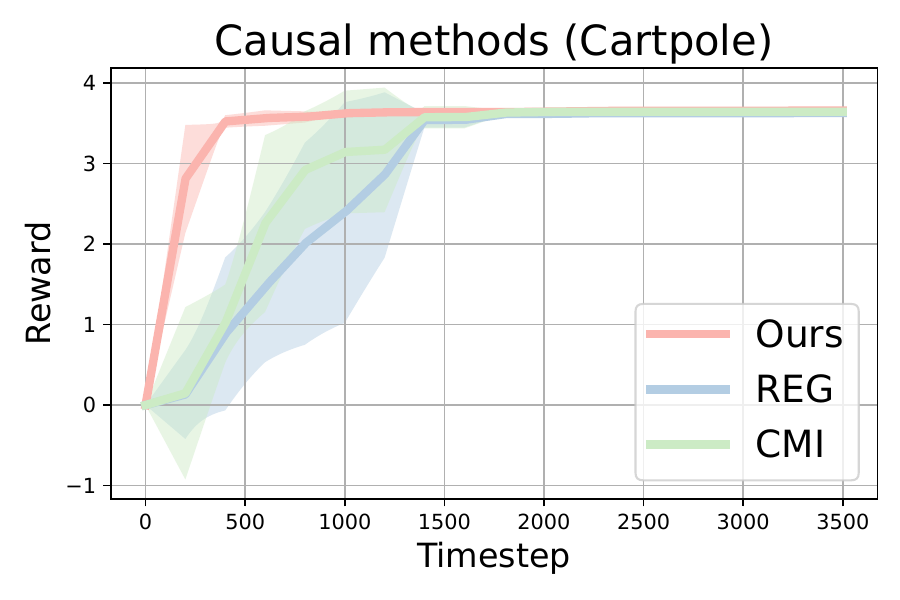}
    \includegraphics[width=0.47\linewidth, height=2.2cm]{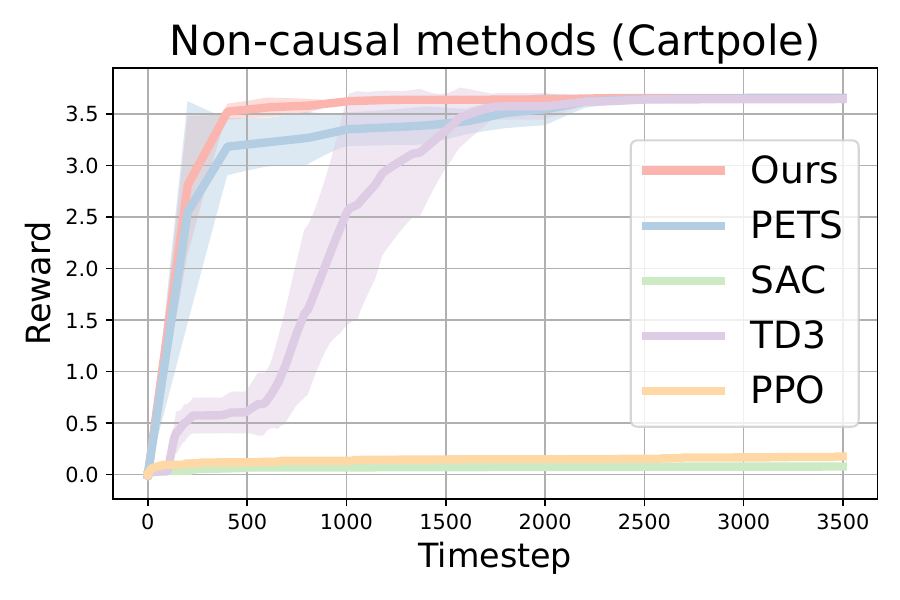}
    \vspace{-8pt}
    \caption{\small{Our model learns faster than other causal, model-based and model-free baselines trained in the Cartpole environment. Statistics (mean $\pm$ std) computed over five runs.}\vspace{-12pt}}
    \label{fig:cartpole-reward}
\end{figure}

\begin{figure} 
    \centering
    \includegraphics[width=0.47\linewidth, height=2.5cm]{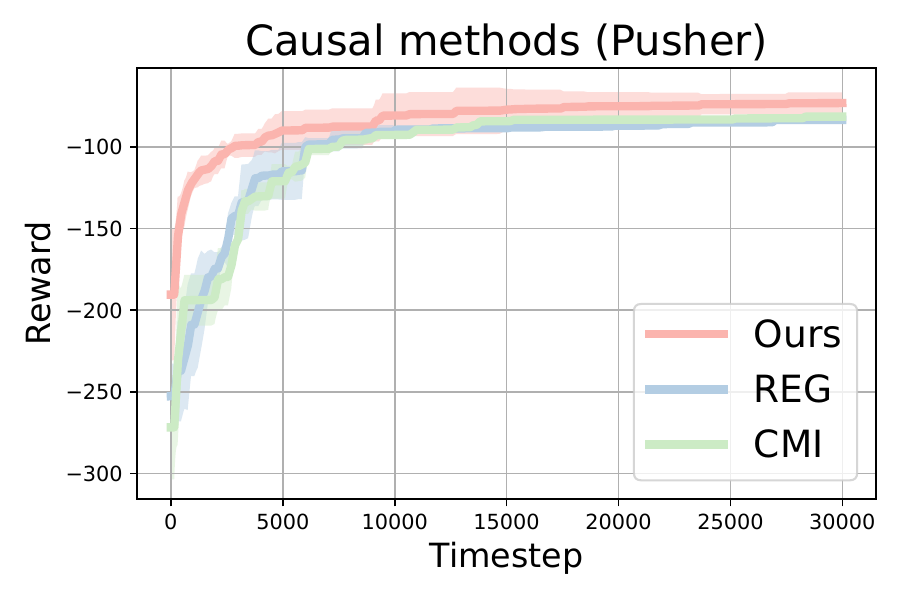}
    \includegraphics[width=0.47\linewidth, height=2.5cm]{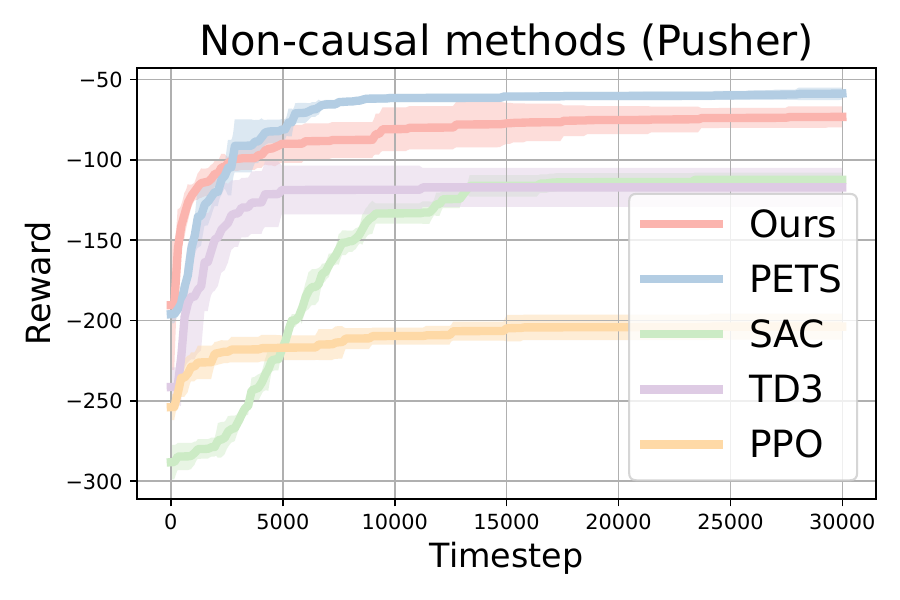}
    \vspace{-8pt}
    \caption{\small{Using lower computational resources, our model achieves similar performance as other causal, model-based and model-free baselines trained in the Pusher environment. Statistics (mean $\pm$ std) computed over five runs (three runs for CMI and Reg).}\vspace{-20pt}}
    \label{fig:pusher-reward}
\end{figure}

Figures \ref{fig:cartpole-reward} and \ref{fig:pusher-reward} illustrate training rewards for Cartpole and Pusher. To emphasize causal methods, we separate the plots into causally-inspired (left) and standard (right) RL approaches. Notably, during the early stages of training, when data is limited, CADY outperforms larger, more powerful baselines. As more data becomes available, it maintains comparable performance or even surpasses it.

\begin{figure}[h] \vspace{-12pt}
\includegraphics[width=0.9\linewidth,height=3cm]{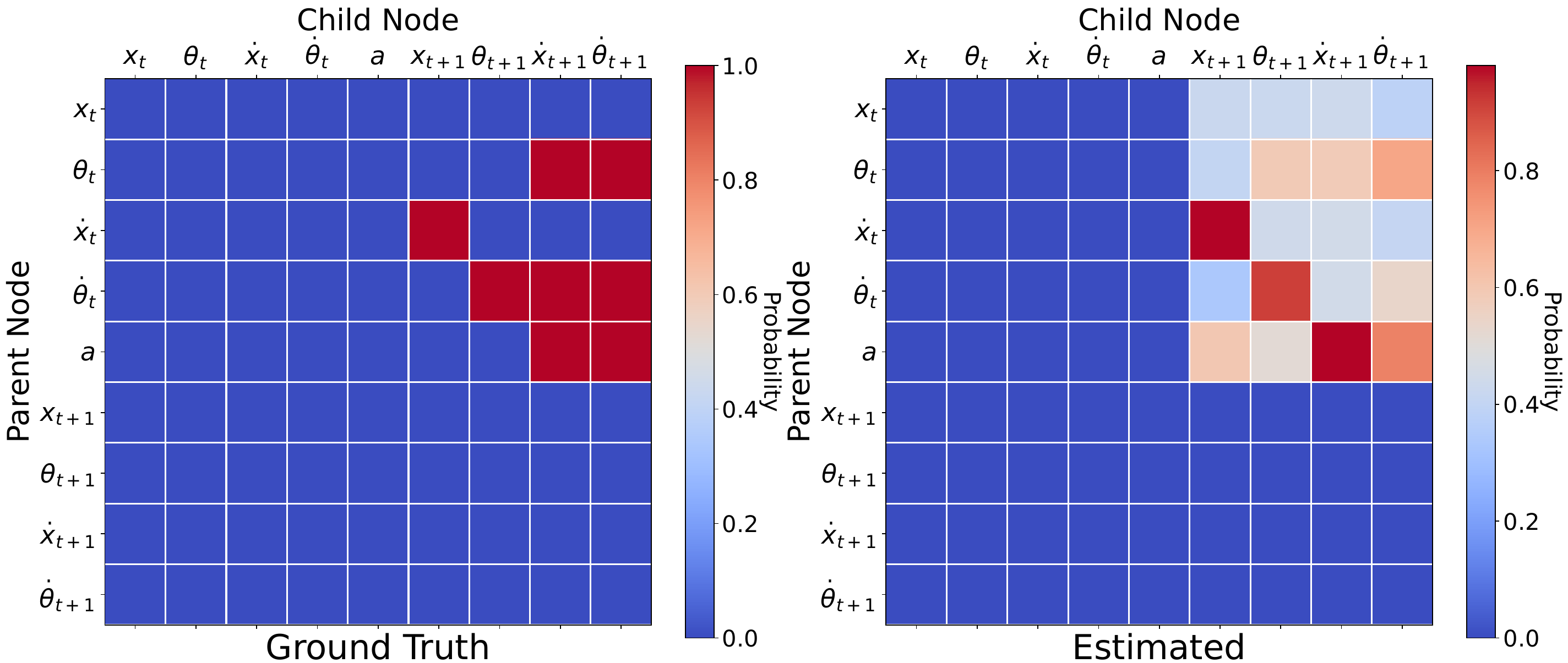} \vspace{-6pt}

\centering
\caption{\small{Estimated Causal Structure Distribution $\mathcal{P}(\mathbf{p})$ for the Cartpole model. \textit{(Left)} Ground truth causal graph. \textit{(Right)} Estimated distribution $\mathcal{P}(\mathbf{p})$. Each cell's color represents the probability $\mathbf{p}$ that an edge exists between two nodes. We can also appreciate the bounded $(n+p)\times n$ search space (edges with zero probability) defining the bipartite structure we strive to estimate.}\vspace{-10pt}} 
\label{fig:causal_structure_adj_matrix}
\end{figure}

Fig. \ref{fig:causal_structure_adj_matrix} presents the \href{https://github.com/facebookresearch/mbrl-lib/blob/3f93cccfc8d635f74e335a2f07aab6e9a48fc021/mbrl/env/cartpole_continuous.py#L60}{ground truth} (left) and estimated (right) causal graph adjacency matrices for the Cartpole environment. These matrices display the probability $p_{ij}$ of a causal link from node $i$ to node $j$ in the true SCM of the environment's dynamics. 
Note that, since we are learning the state delta, some variables, such as $x_t$, do not act as parents for their subsequent values $x_{t+1}$ (the figure does not show the $\Delta$ symbol for ease of visualization). An analogous example for another dynamics model is available in Fig. \ref{fig:causal-graph-example}. 

In general, we can see that the probability mass follows the ground truth, with the exception of the link $\dot{\theta}_t \rightarrow \dot{x}_{t+1}$, which, although non-zero, has a lower probability than expected. We hypothesize that this discrepancy arises from weaker evidence supporting the causal relationship.

We proceed with the best-performing baseline (PETS) for the remaining experiments.

\begin{SCtable} 
\centering
\scriptsize
\setlength{\tabcolsep}{1.5pt}
\renewcommand{\arraystretch}{0.85}
\caption{\small{Computational Requirements. \textit{FLOPS} indicate inference cost, and \textit{parameters} the needed representational capacity.}}\vspace{-6pt} 
\label{table:model_efficiency}
\begin{tabular}{llcccc}
\toprule
 & & \multicolumn{2}{c}{FLOPS $\downarrow$} & \multicolumn{2}{c}{\# Params. $\downarrow$} \\
\cmidrule(lr){3-4} \cmidrule(lr){5-6}
 & & Ours & PETS & Ours & PETS \\
\midrule
Cartpole & & \textbf{235} & 416160 & \textbf{230} & 416048 \\
Pusher & & \textbf{23563} & 470800 & \textbf{23756} & 470240 \\
Jackal & & \textbf{1542} & 409120 & \textbf{1638} & 412060 \\
\bottomrule
\end{tabular}
\end{SCtable}

\vspace{-1mm}
\subsection{Lower Computational Requirements} \label{subsec:results.comp-req}
\vspace{-1mm}
\begin{figure}
    \begin{center}
    \includegraphics[width=0.75\linewidth]{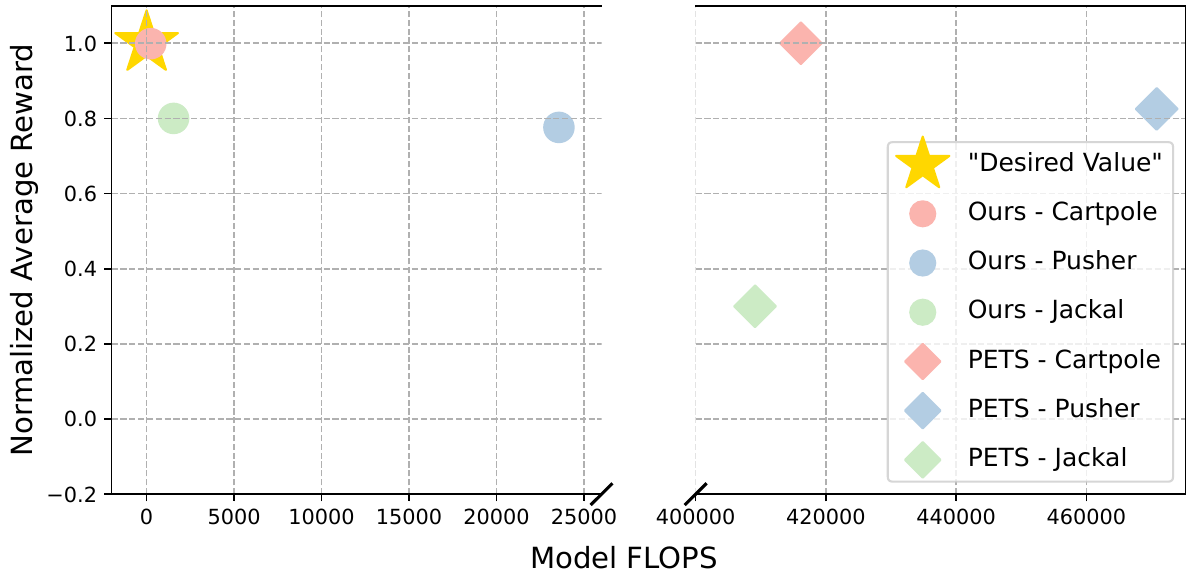}
    \centering
    \caption{\small{Performance as a function of computational requirements. Note the broken $x$-axis for ease of visualization.}}\vspace{-20pt} 
    \label{fig:model_sizes}
    \end{center}
\end{figure}

\begin{figure} 
{
\vspace{-20pt}
    \begin{center}
    \includegraphics[width=0.368\linewidth]{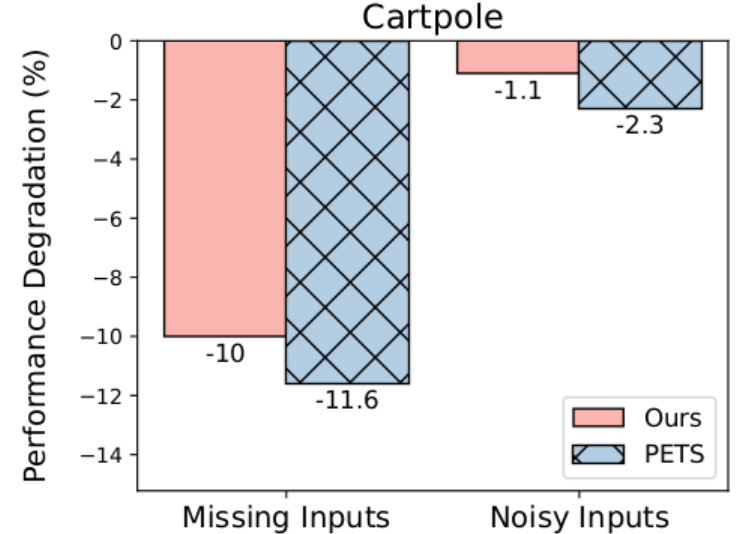}
    \includegraphics[width=0.368\linewidth]{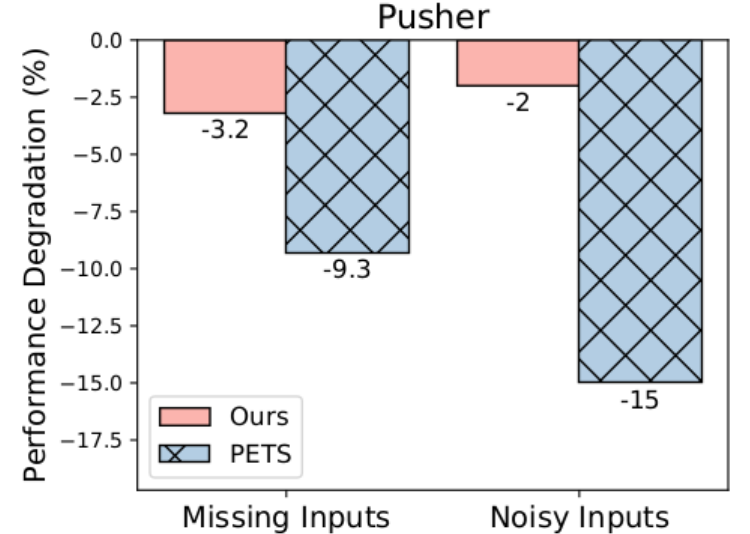}
    \centering
    \caption{\small{Cartpole and Pusher Robustness Evaluation. Performance degradation (\%) when the inputs have missing or noisy features.}} 
    \label{fig:robustness_tests}
    \end{center}
}
\end{figure}

Our method significantly reduces computational requirements compared to the baseline, as shown in Table \ref{table:model_efficiency}. Across all environments, CADY achieves a drastic reduction in both FLOPs (computed with \cite{muckley2021neuralcompression}) and parameter count, with improvements consistently exceeding one to two orders of magnitude. Note that these results consider the overhead of sampling the causal structure at inference time. This efficiency is particularly notable in manipulation tasks, where our model demands nearly \textbf{20$\times$} fewer operations than PETS. Similarly, in the Jackal environment, CADY reduces both FLOPs and parameter count by more than \textbf{250$\times$}. These results illustrate that CADY not only delivers strong performance but also operates with a fraction of the computational cost, making it ideal for real-time execution on resource-constrained platforms.

Fig. \ref{fig:model_sizes} further highlights this efficiency by showing that our models lie closer to the upper-left corner, where optimal performance-cost trade-offs are achieved.

\begin{figure}  \vspace{-15pt} 
    \begin{center}

    \hspace{-1mm}\includegraphics[width=0.93\linewidth, height=0.7in]{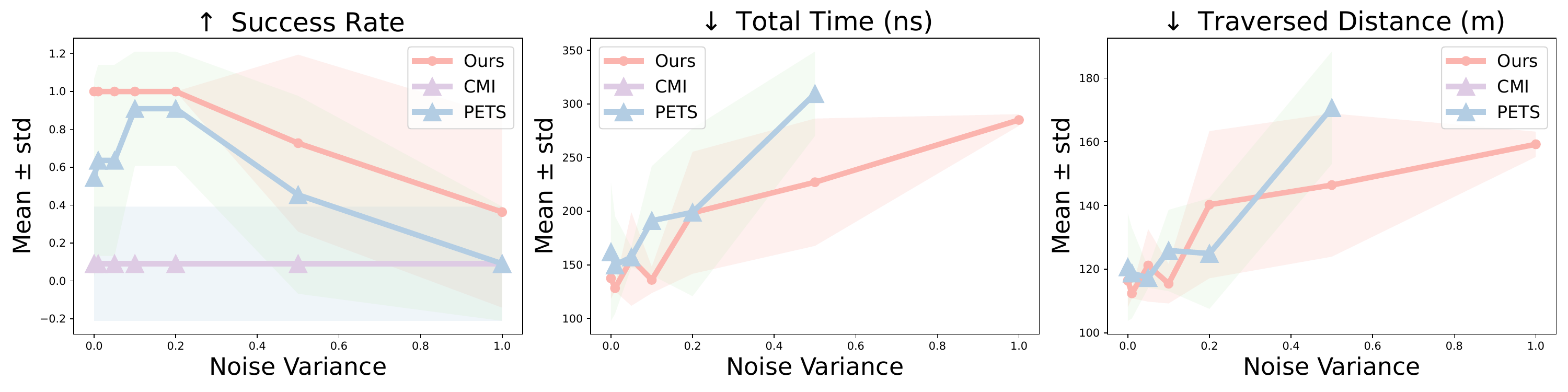}
    \includegraphics[width=0.93\linewidth, height=0.7in]{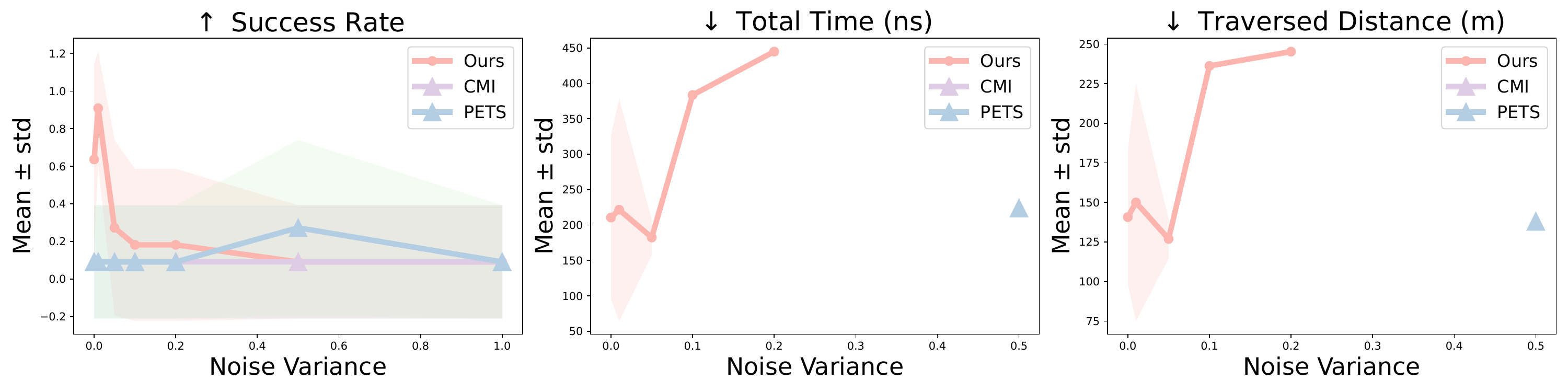}
    \centering
    \caption{\small{Jackal navigation performance with noisy state estimates. 
    Each row shows the results for different missions. Mean $\pm$ std. dev. over 10 trials. Only successful runs are considered for the total time and traveled distance statistics.
    } }
    \label{fig:jackal_experiments}
    \end{center}
\end{figure}

\begin{figure*}[h]  
    \centering
    \includegraphics[width=0.8\linewidth]{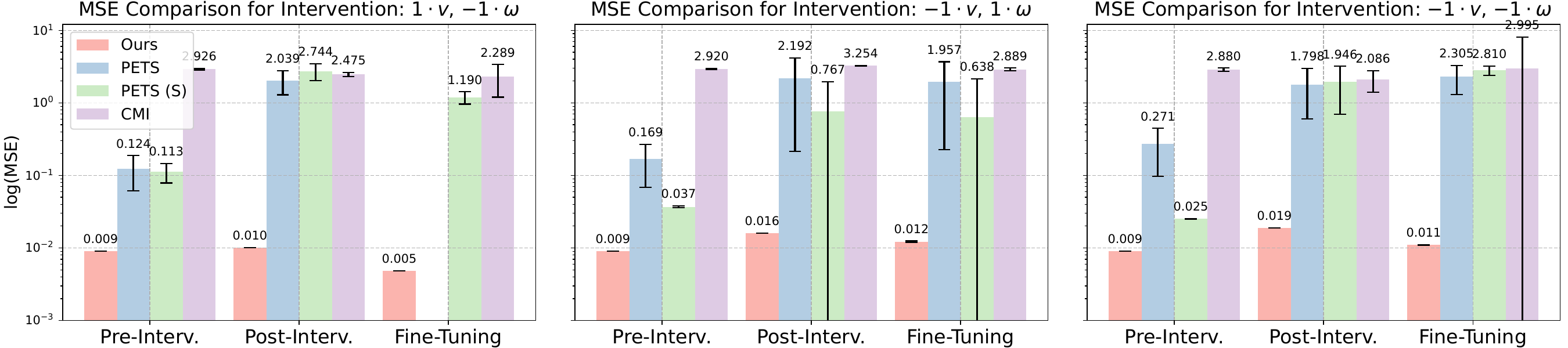}
    \vspace{-6pt}
    \caption{\small{
    Effect of Unexpected Interventions and Online Corrections. Mean Squared Error for one-step prediction with each dynamics model before the intervention (pre-interv.), after the intervention (post-interv.), and after fine-tuning while remaining in the intervened dynamics regime. CMI is included to validate other causally-informed dynamics, while PETS (S) demonstrates that improved fine-tuning performance is not solely due to fewer trainable parameters. Note that the log scale causes variance bars to appear asymmetrical.}\vspace{-16pt}
    }
    \label{fig:interventions}
\end{figure*}

\vspace{-2mm}
\subsection{Increased Robustness to Missing and Noisy Inputs} \label{subsec:results.robustness}
\vspace{-1mm}
Robustness to disruptions in state estimation is crucial for real-world deployment, where sensor failures and noise are common. A model that remains stable under these conditions ensures more reliable planning, reducing the risk of catastrophic failures in downstream tasks. In this section we address \textbf{(Q2)} and evaluate CADY’s ability to handle missing and noisy inputs compared to baseline models.

\textbf{Cartpole and Pusher.} We analyze performance degradation when a state variable ceases to update, simulating scenarios like a \textit{disconnected or malfunctioning sensor}—e.g., due to vibrations during uneven terrain traversal. Specifically, after $10\%$ of a trial's duration, we freeze one state variable and maintain its last observed value for the remainder of the trial. We repeat this for \textit{each} variable in the state vector across five runs. Fig.~\ref{fig:robustness_tests} shows the performance degradation $PD(x) = \frac{x - x_{ref}}{x_{ref}}$,
where $x_{ref}$ is the model's performance in nominal conditions. CADY consistently outperforms the baseline, exhibiting a \textbf{1.6\%} lower degradation in Cartpole and \textbf{6.1\%} lower in Pusher. The performance gap widens in high-dimensional environments like Pusher. We hypothesize that this can be attributed to CADY’s causal structure, which induces sparsity in the latent space, thereby preventing noise propagation and mitigating error accumulation.

Additionally, we evaluate robustness to sensor noise, simulating conditions such as \textit{occlusion} due to outdoor dirt contamination or electromagnetic interference. We inject Gaussian noise $\mathcal{N}(0, 0.05)$ into state observations and run 20 trials. Fig.~\ref{fig:robustness_tests} confirms that CADY features a \textbf{1.2\%} lower degradation in Cartpole and \textbf{13\%} lower in Pusher, reinforcing its superior resilience to real-world uncertainties and highlighting a positive trend in higher dimensions. 

We also study the benefit of the distribution versus a single graph, by repeatedly (10 times) carrying out the tasks. We find that using a fixed graph \textit{reduces} task performance by \(\mathbf{3.2\%}\) on Cartpole and \(\mathbf{5.8\%}\) on the Pusher platform, confirming that marginalizing over plausible structures helps address epistemic uncertainty during forward dynamics prediction.

\begin{table}[] 
\scriptsize
\setlength{\tabcolsep}{3pt}  
\renewcommand{\arraystretch}{0.9}  
\centering
\caption{\small{Results across multiple sampling-distribution variances and state noise levels. Mean $\pm$ std. dev. over 10 trials. Only successful runs are considered for total time and traveled distance statistics. Dashes indicate missing data due to incomplete task.}\vspace{-6pt}} 

\label{tab:jackal-noise-insp-world}
    \begin{tabular}{cl|cccccc}
    \toprule
     &  & \multicolumn{2}{c}{Success Rate $\uparrow$} & \multicolumn{2}{c}{Total Time (s) $\downarrow$} & \multicolumn{2}{c}{Trav. Distance (m) $\downarrow$} \\
     Noise $\sigma^2$ & Model & Mean & Std. Dev. & Mean & Std. Dev. & Mean & Std. Dev. \\
    \bottomrule
    \rowcolor{gray!20} \multicolumn{8}{c}{\textit{Mission 1}} \\
    \multirow{2}{*}{\textbf{0.0}} & \textbf{Ours} & \textbf{1.00} & \textbf{0.00} & 145.87 & 33.98 & 120.65 & 15.65 \\
    \textbf{} & \textbf{PETS} & 0.45 & 0.50 & \textbf{143.00} & \textbf{41.05} & \textbf{116.27} & \textbf{10.94} \\
    \hdashline
    \multirow{2}{*}{\textbf{0.01}} & \textbf{Ours} & \textbf{1.00} & \textbf{0.00} & \textbf{131.68} & \textbf{9.26} & \textbf{114.13} & \textbf{4.49} \\
    \textbf{} & \textbf{PETS} & 0.68 & 0.47 & 154.75 & 42.66 & 119.14 & 11.23 \\
    \hdashline
    \multirow{2}{*}{\textbf{0.05}} & \textbf{Ours} & \textbf{1.00} & \textbf{0.00} & \textbf{143.44} & \textbf{34.60} & 117.68 & 12.68 \\
    \textbf{} & \textbf{PETS} & 0.75 & 0.44 & 154.99 & 29.38 & \textbf{116.79} & \textbf{5.88} \\
    \hdashline
    \multirow{2}{*}{\textbf{0.10}} & \textbf{Ours} & \textbf{0.97} & \textbf{0.16} & \textbf{152.40} & \textbf{37.19} & \textbf{122.06} & \textbf{15.73} \\
    \textbf{} & \textbf{PETS} & 0.88 & 0.33 & 181.71 & 46.55 & 123.72 & 12.15 \\
    \hdashline
    \multirow{2}{*}{\textbf{0.20}} & \textbf{Ours} & \textbf{1.00} & \textbf{0.00} & 183.11 & 65.68 & 133.34 & 26.13 \\
    \textbf{} & \textbf{PETS} & 0.93 & 0.27 & \textbf{170.18} & \textbf{48.90} & \textbf{119.32} & \textbf{10.94} \\
    \hdashline
    \multirow{2}{*}{\textbf{0.50}} & \textbf{Ours} & \textbf{0.72} & \textbf{0.45} & \textbf{237.05} & \textbf{69.13} & \textbf{149.97} & \textbf{24.68} \\
    \textbf{} & \textbf{PETS} & 0.40 & 0.50 & 289.11 & 56.09 & 163.90 & 22.30 \\
    \hdashline
    \multirow{2}{*}{\textbf{1.0}} & \textbf{Ours} & \textbf{0.15} & \textbf{0.36} & \textbf{283.11} & \textbf{22.82} & \textbf{160.15} & \textbf{9.77} \\
    \textbf{} & \textbf{PETS} & 0.00 & 0.00 & - & - & - & - \\
    \bottomrule
    \rowcolor{gray!20} \multicolumn{8}{c}{\textit{Mission 2}} \\
    \multirow{2}{*}{\textbf{0.0}} & \textbf{Ours} & \textbf{0.68} & \textbf{0.47} & \textbf{218.30} & \textbf{128.10} & \textbf{146.31} & \textbf{55.92} \\
    \textbf{} & \textbf{PETS} & 0.00 & 0.00 & - & - & - & - \\
    \hdashline
    \multirow{2}{*}{\textbf{0.01}} & \textbf{Ours} & \textbf{0.65} & \textbf{0.48} & \textbf{228.47} & \textbf{139.21} & \textbf{145.85} & \textbf{55.98} \\
    \textbf{} & \textbf{PETS} & 0.00 & 0.00 & - & - & - & - \\
    \hdashline
    \multirow{2}{*}{\textbf{0.05}} & \textbf{Ours} & \textbf{0.28} & \textbf{0.45} & \textbf{189.16} & \textbf{66.60} & \textbf{128.77} & \textbf{22.50} \\
    \textbf{} & \textbf{PETS} & 0.00 & 0.00 & - & - & - & - \\
    \hdashline
    \multirow{2}{*}{\textbf{0.10}} & \textbf{Ours} & \textbf{0.17} & \textbf{0.38} & \textbf{275.12} & \textbf{139.38} & \textbf{173.95} & \textbf{67.78} \\
    \textbf{} & \textbf{PETS} & 0.00 & 0.00 & - & - & - & - \\
    \hdashline
    \multirow{2}{*}{\textbf{0.20}} & \textbf{Ours} & 0.05 & 0.22 & 471.54 & 37.85 & 238.77 & 9.25 \\
    \textbf{} & \textbf{PETS} & \textbf{0.07} & \textbf{0.27} & \textbf{176.64} & \textbf{19.07} & \textbf{129.45} & \textbf{12.54} \\
    \hdashline
    \multirow{2}{*}{\textbf{0.50}} & \textbf{Ours} & 0.03 & 0.16 & 387.82 & - & 193.95 & - \\
    \textbf{} & \textbf{PETS} & \textbf{0.15} & \textbf{0.36} & \textbf{277.98} & \textbf{133.41} & \textbf{162.35} & \textbf{45.66} \\
    \hdashline
    \multirow{2}{*}{\textbf{1.0}} & \textbf{Ours} & 0.00 & 0.00 & - & - & - & - \\
    \textbf{} & \textbf{PETS} & 0.00 & 0.00 & - & - & - & - \\
    \bottomrule
    \end{tabular}\vspace{-16pt}
\end{table}

\textbf{Jackal.} We extend the robustness evaluation to Jackal in the Inspection World simulator. To simulate real-world localization challenges—such as GPS degradation in forests or urban canyons—we inject noise into state estimates, with variance $\sigma^2 \in \{0.01, 0.05, 0.1, 0.2, 0.5, 1\}$. Fig.~\ref{fig:jackal_experiments} shows that CADY outperforms the baseline across all noise levels and missions. In Mission 1, the robot achieved \textbf{85.7\%} success rate (vs. 55.7\%), \textbf{18.1\%} lower mission time and \textbf{4\%} lower traversed distance, showing a significantly higher efficiency. We pick Mission 2 given its challenging nature, as the robot needs to climb a steep mountain where deviations from the path can make it slip. Although we see lower results in this hard task, we still achieved \textbf{27\%} success rate (vs. 2.8\%). For videos of CADY completing the missions, visit: \href{https://youtu.be/v-WnfQF0hyI}{\underline{Clip 1}} and \href{https://youtu.be/Du3ASfPx6-I}{\underline{Clip 2}}. Overall, across missions, our method achieved \textbf{27.1\%} higher success rate and \textbf{2.1\%} lower mission completion time. We also perform a sensitivity analysis of MPPI's sampling distribution variance, testing values of 0.01, 0.1, 0.4, and 0.8 over 10 repetitions in the two missions. This aims to show that the robustness stems from CADY and not from arduously tuned hyperparameters. Results are summarized in Table \ref{tab:jackal-noise-insp-world}, where CADY again shows a better performance.

\vspace{-3mm}
\subsection{Effects of Unmodeled Interventions} \label{subsec:results.adaptation}
\vspace{-1mm}

Next, we address \textbf{(Q3)}:  
To further evaluate the properties of causal masking in CADY, we introduce interventions in the form of manipulated controls to simulate scenarios where the dynamics change (e.g., terrain with varying friction) or the robot’s sensors provide inaccurate measurements. These interventions are represented as gains applied to the \textit{simulator's} control inputs, but both the value of these gains and the timing of their application are unknown to the dynamics models. Specifically, the system operates under nominal dynamics for the first 250 steps. At this point, an intervention alters the environment’s dynamics, introducing a controlled perturbation. The robot continues to operate under these new conditions for another 250 steps, after which each dynamics model undergoes fine-tuning using the most recent trajectory data. This setup enables us to evaluate how effectively the models adapt to sudden changes and whether CADY mitigates the impact of these interventions. A more robust dynamics model directly translates to improved model-based planning.  

Fig. \ref{fig:interventions} compares the learned Jackal dynamics across multiple intervention scenarios. The gains applied to the linear $v$ and angular $\omega$ velocities are indicated above each subplot. Each subplot presents the one-step prediction error for each model pre-intervention, immediately after the intervention (post-interv.), and following fine-tuning. Notably, the new dynamics regime remains active after fine-tuning. The results clearly demonstrate that CADY significantly outperforms all baselines, despite being trained on the same dataset and under identical conditions (described in Section \ref{subsec:eval-setup}).  

Importantly, the minimal performance degradation observed in the post-intervention phase highlights the advantages of our causally informed model. For instance, under the first intervention schedule, which affects only the applied angular velocity $\omega$, the performance decrease is lower ($0.010$) compared to the second intervention schedule, which alters the applied linear velocity $v$ ($0.016$). This difference arises from Jackal's differential drive mechanism: while $\omega$ influences only a single predicted variable ($\theta_{t+1}$), $v$ impacts two ($x_{t+1}$ and $y_{t+1}$), as depicted in the causal graph in Fig. \ref{fig:causal-graph-example}. This signals that the model is able to reduce the influence of non-causal variables, which is not the case in other learning-based methods where all inputs are fed to the model independent of any causal considerations. On the fine-tuning side, we observe that updating the weights of the model with a small dataset incorporating recent measurements enhances its ability to adapt to sudden changes. 

\vspace{-3mm}
\subsection{Field Demonstrations} \label{subsec:results.field-exp}
\vspace{-1mm}
We use Jackal to demonstrate CADY's zero-shot transfer from simulation to reality. All our code runs onboard the robot with an NVIDIA Jetson Orin, a Velodyne 16 LiDAR and an 3DM-GX5-AHRS IMU.
See details in our \textbf{video}: \url{https://youtu.be/X6k5t7OOnNc}.


\vspace{-3mm}
\section{Conclusions}
\vspace{-2mm}

We show that learning the robot's dynamics by considering a distribution over causal structures, instead of a single causal structure, gives downstream planners improved robustness. We have introduced a model for the distribution of causal structures and a method based on FA to estimate the distribution's parameters. We also present CADY, a dynamics model based on a probabilistic encoder-multidecoder architecture, that informs its latent space via the learned distribution of causal structures. CADY features lower computational requirements and better prediction performance under common, realistic and challenging conditions. We have validated our method with manipulators and mobile-robots navigating on uneven terrain, in both simulation and the real-world.
 
\bibliographystyle{unsrt}
\bibliography{references}

\end{document}